\documentclass[letterpaper, 10 pt, conference]{ieeeconf}
\IEEEoverridecommandlockouts
\usepackage{booktabs}
\usepackage{balance}
\usepackage{cite}
\usepackage{amsmath,amssymb,amsfonts}
\usepackage{algorithm}
\usepackage{algpseudocode}
\usepackage{graphicx}
\usepackage{textcomp}
\usepackage{xcolor}
\usepackage{verbatim}
\usepackage{graphicx}
\usepackage{multirow}
\usepackage{mathrsfs}
\usepackage{hyperref}

\IEEEoverridecommandlockouts
\begin{document}

\title{A Topology-Aware Spatiotemporal Handover Framework for Continuous Multi-UAV Tracking}
\author{Jianlin~Ye,~Christos~Kyrkou and Panayiotis~Kolios% <-this % stops a space
\thanks{The authors are with the KIOS Research and Innovation Centre of Excellence (KIOS CoE), and University of Cyprus, Nicosia, 1678, Cyprus. {\tt\small \{ye.jianlin, kyrkou.christos, pkolios\}@ucy.ac.cy}}
}
\maketitle

\begin{abstract}
The integration of Unmanned Aerial Vehicles (UAVs) into Intelligent Transportation Systems (ITS) offers synoptic visibility for traffic monitoring, yet scalable deployment is hindered by trajectory fragmentation, where vehicle identity persistence is lost across multi-UAV Fields of View (FOV). While state-of-the-art frameworks excel in optimizing local trajectory extraction and stability for single-drone imagery, they often function as isolated data silos that generate disjointed trajectories, thereby precluding network-level analysis such as Origin-Destination estimation. This paper presents a real-time Multi-Camera Multi-Vehicle Tracking (MCMT) system designed to handle global identity persistence. Addressing the visual ambiguity and computational cost of appearance-based Re-Identification (Re-ID) in nadir views, we introduce a lightweight Topology-Based Spatiotemporal Handover mechanism. We implement a high-throughput parallel pipeline leveraging YOLO11 and ByteTrack to process concurrent 4K streams. Our core contribution is a deterministic queue-based matching algorithm that utilizes geometric overlaps and virtual lane discretization to predictively manage identity handover via FIFO queues. Experimental results on complex urban environments, including intersections and merging traffic, demonstrate a Handover Success Rate (HOSR) of 99.8\% in continuous traffic flows, significantly outperforming Re-ID baselines (74.1\%) while validating edge deployment feasibility. The source code is available at \url{https://github.com/JYe9/multi-camera-multi-vehicle-tracking-system}.
\end{abstract}
\section{Introduction} 
\label{sec:Introduction}

Intelligent Transportation Systems (ITS) traditionally depended on spatially sparse fixed infrastructure, which captures discrete spot speeds but remains blind to continuous flow dynamics evolving between sensor nodes~\cite{buch2011review}. To bridge this gap, Unmanned Aerial Vehicles (UAVs) are increasingly utilized for traffic monitoring~\cite{kanistras2013survey,barmpounakis2016unmanned,bisio2022systematic}. UAVs provide a synoptic perspective that mitigates occlusion~\cite{highd2018}, enabling ad-hoc monitoring of congestion without the capital expenditure of permanent installations.

However, widespread UAV deployment introduces distinct constraints. To capture vehicles with sufficient spatial resolution, commercial quadcopters must operate at altitudes that inherently restrict their Field of View (FOV). Consequently, monitoring operationally relevant corridors requires a linear topology of multiple UAVs, creating the critical problem of trajectory fragmentation. In standard pipelines, individual UAVs function as autonomous observers, generating isolated local identities for the same vehicle as it traverses sequential views. Lacking a unified identity, crucial macroscopic metrics like average travel time and long-distance lane-changing frequency cannot be accurately derived. This fragmentation yields disconnected tracklets rather than global trajectories, rendering the data insufficient for network-level metrics like Origin-Destination estimation.

To associate these fragmented tracklets, existing solutions often rely on Appearance-based Re-Identification (Re-ID)~\cite{tang2019cityflow}. While effective in ground-level surveillance, visual Re-ID frequently fails in aerial monitoring due to nadir-view ambiguity, where vehicles of similar models appear indistinguishable~\cite{du2019visdrone,xia2018dota}. Moreover, the high computational cost of extracting visual embeddings often precludes real-time execution on resource-constrained onboard hardware. This reliance on visual features often overlooks the logical connectivity between sensors, leaving the challenge of real-time, deterministic identity persistence largely unaddressed. 

\begin{figure*}
    \centering
    \includegraphics[width=1.0\textwidth]{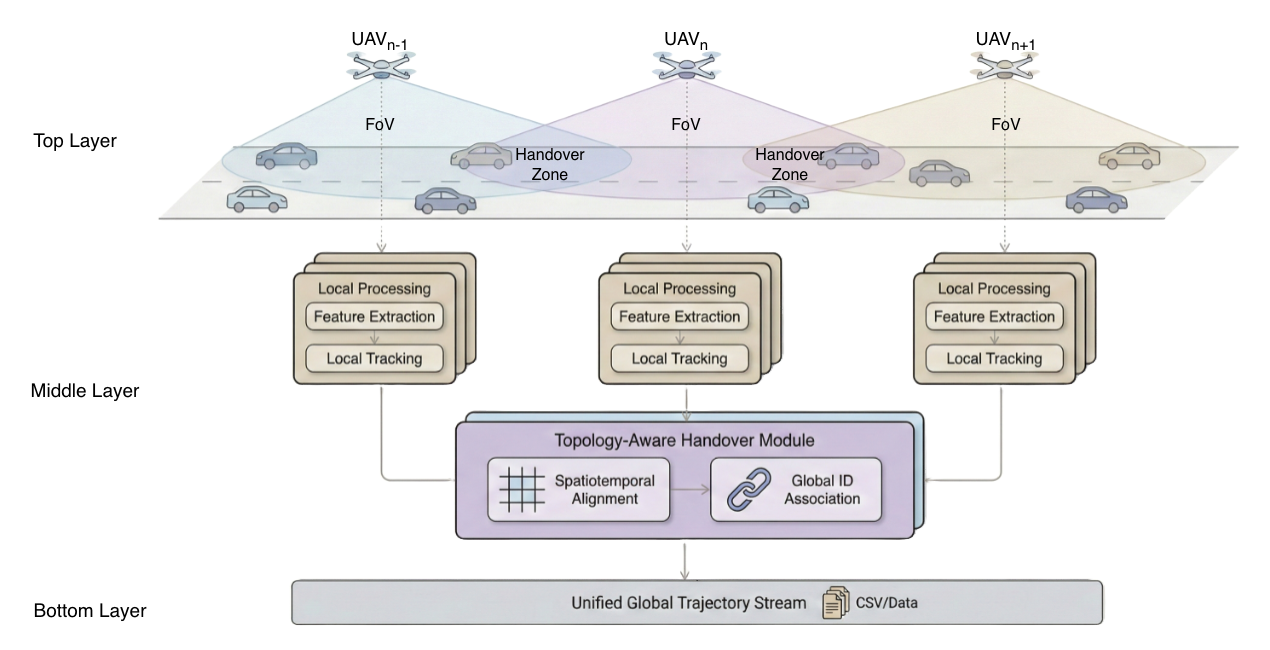}
    \caption{The proposed Unified Aerial Surveillance (UAS) framework for multi-UAV urban vehicle tracking. The pipeline is structured into three hierarchical layers: (1) Top Layer (Physical Scenario): Multiple UAVs collaboratively monitor the urban traffic flow, where adjacent Fields of View (FOVs) overlap to form Handover Zones for seamless vehicle transition; (2) Middle Layer (Processing Core): Individual UAVs perform local feature extraction and tracking, feeding into the core Topology-Aware Handover Module for spatiotemporal alignment and global ID association; (3) Bottom Layer (Data Output): The system synthesizes fragmented local data into a unified, continuous global trajectory stream.}
    \label{fig:pipeline}
\end{figure*}

To overcome these limitations, this paper presents a unified aerial surveillance system designed to enforce global identity persistence. We propose a shift from fragile visual matching to a robust topology-based handover mechanism, leveraging the constraint that traffic flow follows a predictable spatiotemporal continuum. Specifically, we treat the multi-UAV chain not as a set of independent sensors, but as a logically connected pipeline where vehicle exits and entries are causally linked events, allowing for deterministic association even in the absence of distinctive visual features. The primary contributions of this work are three-fold:

\begin{itemize}
    \item We introduce a deterministic Spatiotemporal Queue-based Matching algorithm. By utilizing geometrically overlapping polygons and virtual lane discretization, we populate directional First-In-First-Out (FIFO) queues that stitch fragmented tracklets into cohesive global trajectories without relying on visual features.
    \item We propose a metric-aware kinematic state estimator that transforms raw pixel tracklets into calibrated velocity and heading profiles, enabling precise lane-level traffic flow analysis.
    \item We validate the system's robustness in urban environments featuring intersections and merging traffic, proving topological constraints offer a scalable and computationally efficient alternative to appearance-based methods for continuous wide-area tracking.
\end{itemize}

The proposed hierarchical architecture is illustrated in Fig.~\ref{fig:pipeline}. The remainder of this paper is organized as follows. Section \ref{sec:Related_Work} reviews related work, Sec. \ref{sec:Problem} formulates the tracking problem, and Sec. \ref{sec:Approach} details the proposed system. Finally, Sec. \ref{sec:Evaluation} presents experimental results, and Sec. \ref{sec:Conclusion} concludes the paper.
\section{Related Work}
\label{sec:Related_Work}

\subsection{Deep Learning in Aerial Object Detection}
Aerial imagery constitutes an adversarial domain, primarily due to extreme scale variation where objects occupy minimal frame area, compounded by dynamic occlusion from vegetation~\cite{du2019visdrone,xia2018dota}. While the You Only Look Once (YOLO) family~\cite{redmon2016yolo} has established the standard for real-time inference, early anchor-based iterations often failed to resolve dense clusters of small vehicles. A significant paradigm shift occurred with the introduction of anchor-free mechanisms in YOLOX~\cite{ge2021yolox}, which decoupled the prediction head to improve localization accuracy. Contemporary architectures, such as the YOLO11 framework~\cite{jocher2023ultralytics}, have further evolved to integrate attention mechanisms like Cross-Stage Partial Spatial Attention. These refinements are critical for preserving semantic features of small objects at high resolutions. However, differentiating visually similar classes in nadir views remains a persistent challenge that detection models cannot resolve without temporal or topological context.

\begin{figure*}[t]
    \centering
    \includegraphics[width=1.0\textwidth]{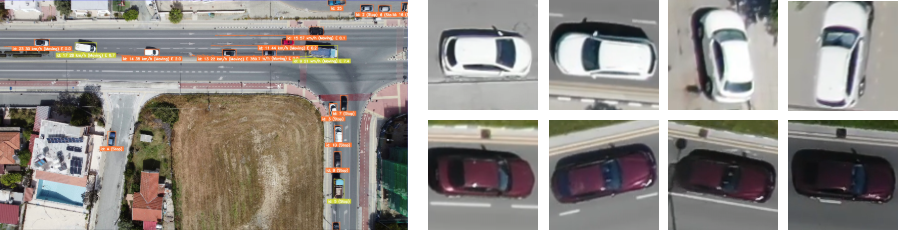}
    \caption{Single-UAV processing outputs and key appearance challenges in nadir-view aerial traffic monitoring. \textbf{Left:} a representative UAV footprint where the system performs detection, multi-object tracking, and per-vehicle attribute estimation (e.g., velocity and vehicle type). \textbf{Right (top):} nadir-view ambiguity where multiple visually similar vehicles (e.g., white cars) are difficult to distinguish. \textbf{Right (bottom):} illumination variation where the same vehicle can exhibit markedly different appearances under cloud cover, building shadows, or direct sunlight.}
    \label{fig:challenges}
\end{figure*}

\subsection{Tracking and Multi-Camera Association}
\subsubsection{Local Association}
To bridge detection and trajectory generation, recent advances in Multi-Object Tracking (MOT) have moved beyond simple Kalman filtering. Methods such as ByteTrack~\cite{zhang2022bytetrack} and OC-SORT~\cite{cao2023ocsort} maximize detection association by utilizing low-confidence bounding boxes, a strategy proven to reduce track fragmentation in occlusion-heavy scenarios. This capability is particularly relevant for aerial surveillance, where vehicle confidence scores fluctuate frequently due to changing viewing angles and environmental shadows.

\subsubsection{Multi-Camera Handover}
Transitioning from single-view to Multi-Camera Tracking (MCT) necessitates robust mechanisms to associate tracklets across disjoint views. The dominant paradigm, Appearance-based Re-Identification (Re-ID)~\cite{tang2019cityflow}, extracts high-dimensional embeddings to match vehicles via cosine similarity. While effective for distinctive objects, Re-ID faces severe limitations in aerial highway monitoring. The nadir-view ambiguity renders visual features such as grilles and wheels invisible, leading to high inter-class similarity among standard vehicles. Furthermore, the quadratic computational complexity of matching matrices often precludes real-time execution on edge devices.

Consequently, topology-based methods have emerged as a scalable alternative by leveraging physical environmental constraints. By modeling the connectivity between camera nodes, the search space for association is reduced to physically plausible transitions~\cite{hsu2019multi,hsu2021multi}. While prior works employed probabilistic Kalman filters to predict arrival times in unmonitored zones, such predictions are sensitive to non-linear speed variations. Recent approaches suggest that enforcing deterministic geometric constraints, such as overlapping fields of view, can eliminate the stochastic errors inherent in purely probabilistic or appearance-based matching.

\subsection{Cooperative Multi-UAV Systems}
Beyond vision-centric algorithms, multi-UAV coordination has been extensively studied from a systems perspective. Research utilizing distributed Model Predictive Control (MPC)~\cite{wolfe2020distributed} or mobile network infrastructure~\cite{zahinos2022cooperative} focuses on maintaining physical formation and coverage. In parallel, advanced photogrammetry frameworks like those by Fonod et al.~\cite{fonod2025advanced} have achieved high precision in georeferencing and track stabilization. However, these works typically treat the multi-UAV setup as a means to expand spatial coverage rather than as a logically connected network. This limitation often results in fragmented identities for vehicles traversing the entire corridor, highlighting the need for a unified topological logic to stitch high-quality local tracklets into cohesive, network-level global trajectories.
\section{Problem Formulation}
\label{sec:Problem}

In this section, we formulate the Multi-Camera Multi-Vehicle Tracking (MCMT) task as a global identity association problem within a decentralized sensor network. We establish the mathematical notation for the network topology, the vehicle state space, and the spatiotemporal constraints governing the handover mechanism.

\subsection{Network Topology and Field of View}
Consider a set of $N$ aerial sensor nodes, denoted as $\mathcal{C} = \{C_1, C_2, \dots, C_N\}$. Each node $C_i$ possesses a local Field of View (FOV) whose geometric projection onto the ground plane is denoted by $\Omega_i \subset \mathbb{R}^2$. The sensor network is modeled as a directed graph $\mathcal{G} = (\mathcal{C}, \mathcal{E})$, where an edge $e_{i,j} \in \mathcal{E}$ exists if and only if the ground projections of node $C_i$ and $C_j$ are spatially overlapping, permitting sequential vehicle transition.

To enable deterministic tracking handover, we enforce a geometric overlap constraint. For any connected pair $(C_i, C_{i+1})$, there exists a non-empty intersection region:
\begin{equation}
    \mathcal{R}_{overlap}^{i, i+1} = \Omega_i \cap \Omega_{i+1} \neq \emptyset
\end{equation}
Within this shared spatial domain $\mathcal{R}_{overlap}$, a physical target is theoretically observable by both sensors simultaneously, subject to synchronization latency and perspective distortion.

\subsection{Local Tracklet Definition}
Let the state of a vehicle observed by sensor $C_i$ at time $t$ be represented by a vector $\mathbf{x}_t^i = [u, v, \dot{u}, \dot{v}, \theta]^\top$, comprising the centroid coordinates $(u,v)$ in the metric domain, velocity components $(\dot{u}, \dot{v})$, and heading angle $\theta$. A \textit{local tracklet} $\mathcal{T}^L$ is defined as a temporal sequence of such states associated with a unique local identifier $id_L$:
\begin{equation}
    \mathcal{T}^L_i = \{ (\mathbf{x}_t^i, t) \mid t \in [t_{start}, t_{end}] \}
\end{equation}
Due to the disjoint nature of standard local processing, a single vehicle traversing the network generates a set of fragmented tracklets $\{\mathcal{T}^L_1, \mathcal{T}^L_2, \dots, \mathcal{T}^L_N\}$ across different nodes, each with an independent local identifier.

\subsection{The Spatiotemporal Handover Problem}
The objective of the proposed system is to derive a local minimum-cost assignment function $\Phi$ that associates a terminating tracklet in the upstream node $C_i$ with an initiating tracklet in the downstream node $C_{i+1}$.

Let $\mathcal{S}_{exit}^i$ denote the set of tracklets exiting $\Omega_i$ through $\mathcal{R}_{overlap}$, and $\mathcal{S}_{entry}^{i+1}$ denote the set of tracklets entering $\Omega_{i+1}$ within the same region. The handover problem is formulated as finding the optimal association pair $(\mathcal{T}_a \in \mathcal{S}_{exit}^i, \mathcal{T}_b \in \mathcal{S}_{entry}^{i+1})$ that minimizes the spatiotemporal association cost $\mathcal{J}$:
\begin{equation}
    \Phi(\mathcal{T}_a) = \underset{\mathcal{T}_b \in \mathcal{S}_{entry}^{i+1}}{\arg\min} \mathcal{J}(\mathcal{T}_a, \mathcal{T}_b)
\end{equation}
subject to the following consistency constraints:
\begin{align}
    \text{Spatial:} & \quad || \mathbf{p}_{last}(\mathcal{T}_a) - \mathbf{p}_{first}(\mathcal{T}_b) ||_2 < \epsilon_{dist} \\
    \text{Temporal:} & \quad | t_{end}(\mathcal{T}_a) - t_{start}(\mathcal{T}_b) | < \epsilon_{time} \\
    \text{Directional:} & \quad \cos(\theta_a - \theta_b) > \gamma_{dir}
\end{align}
where $\mathbf{p}$ denotes the position vector, and $\gamma_{dir}$ represents the minimum cosine similarity for heading alignment. Unlike traditional probabilistic matching, our system solves Eq. (4) using a lateral-aware spatiotemporal buffering mechanism, which enforces these constraints through temporal search windows and spatial metadata matching.
\section{Proposed Approach}
\label{sec:Approach}

The proposed framework addresses the MCMT problem through a hierarchical pipeline. As formulated in Section~\ref{sec:Problem}, the system decomposes the global association task into three coupled stages: (1) Local Perception and Tracklet Generation, (2) Kinematic State Estimation, and (3) Topology-Aware Global Handover.

\subsection{Perception and Local Tracklet Generation}
The foundation of the pipeline is the extraction of local tracklets $\mathcal{T}^L$ from raw aerial imagery. We deploy the YOLO11 architecture, fine-tuned on aerial domains to handle the scale variation inherent in $4K$ inputs~\cite{jocher2023ultralytics}. To minimize false negatives for small objects, the inference resolution is set to $1280$ pixels with a confidence threshold of $\tau_{conf}=0.25$. 

For temporal association within a single node $\Omega_i$, we employ the ByteTrack algorithm~\cite{zhang2022bytetrack}. Unlike standard Kalman-IoU approaches that discard low-confidence detections, ByteTrack's two-stage association logic is critical for maintaining tracklet continuity ($\mathcal{T}^L$) when vehicles traverse areas of high occlusion. This robustness ensures that the tracklet state $\mathbf{x}_t$ remains valid up to the boundary of the overlap region $\mathcal{R}_{ovlp}$, which is a prerequisite for the subsequent handover logic.

\subsection{Topology-Aware Handover Mechanism}
This module solves the optimization problem defined in Eq. (4) by converting the spatiotemporal constraints into a deterministic buffering process. The mechanism relies on the precise geometric calibration of the sensor network topology $\mathcal{G}$ and utilizes coordinate-level metadata to resolve complex multi-vehicle interactions, including parallel overtaking maneuvers. The complete logic for the handover process is formalized in Alg.~\ref{alg:handover_logic}.

\subsubsection{Geometric Overlap Calibration}
Contrary to naive rectangular bounding boxes, we define the overlap region $\mathcal{R}_{ovlp}$ as a non-convex polygon strictly calibrated to the road surface geometry. This polygon accounts for UAV yaw and camera perspective projection. For a vehicle tracklet $\mathcal{T}^L$, its interaction with the boundary is determined by checking if its centroid $\mathbf{p}_t \in \mathcal{R}_{ovlp}$. This geometric constraint serves as the spatial trigger for the handover event.

\subsubsection{Directional Spatial Partitioning}
To satisfy the directional constraint (Eq. 6), we implement a macro-level spatial discretization strategy. Assuming a road-aligned coordinate system, the roadway is logically bisected based on the primary traffic flow directions. We define a partitioning function $\mathcal{L}(\mathbf{x}_t)$ that maps a vehicle to a \textit{Directional Zone}:
\begin{equation}
    \mathcal{L}(\mathbf{x}_t) = 
    \begin{cases} 
    Z_{upper} & \text{if } y < Y_{split} \\
    Z_{lower} & \text{if } y > Y_{split}
    \end{cases}
\end{equation}
where $Z_{upper}$ and $Z_{lower}$ correspond to opposing traffic streams (e.g., Eastbound vs. Westbound). This partitioning acts as a coarse-grained filter, creating isolated logic channels for opposing traffic streams. This ensures that a vehicle exiting Node $i$ in the Eastbound zone can never be erroneously matched with a vehicle entering Node $i+1$ in the Westbound zone, regardless of visual similarity.

\subsubsection{Lateral-Aware Spatial Matching}
\label{sec:Matching_logic}
The handover engine maintains a set of directional buffers $\mathcal{B}_{i \to j}^Z$ for each zone $Z$. Unlike strict FIFO queues that rely solely on temporal order, our system employs Coordinate Metadata Matching to handle simultaneous handovers (e.g., parallel driving or overtaking) within the same zone.

\begin{itemize}
    \item Push Operation (Exit Event): When a tracklet $\mathcal{T}_a$ enters $\mathcal{R}_{ovlp}$ at Node $i$, it is pushed to the corresponding directional buffer $\mathcal{B}_{i \to i+1}^Z$. Crucially, the system stores not just the $gID$, but the full state metadata $\mathbf{M}_a = \{gID, t_{exit}, y_{rel}\}$, where $y_{rel}$ is the lateral position normalized to the road width.
    \item Match \& Pop Operation (Entry Event): When a new tracklet $\mathcal{T}_b$ is initialized within $\mathcal{R}_{ovlp}$ at Node $i+1$, the system queries the upstream buffer $\mathcal{B}_{i \to i+1}^Z$. To accommodate potential synchronization latencies or physical gaps between FOVs, the search scope extends over a temporal window $\Delta t$. The system retrieves the optimal candidate $\mathcal{T}_a$ by minimizing the lateral spatial displacement subject to this temporal constraint:
    \begin{equation}
        \mathcal{T}_{match} = \underset{\mathcal{T}_a \in \mathcal{B}, |t_a - t_b| < \Delta t}{\arg\min} | y_{rel}(\mathcal{T}_a) - y_{rel}(\mathcal{T}_b) |
    \end{equation}
    The inclusion of the temporal bound $|t_a - t_b| < \Delta t$ inherently enables a "Soft Handover" capability, allowing the system to recover global identities even when vehicles traverse unmonitored blind spots or imperfectly calibrated overlap regions. To further filter erroneous associations from non-handover traffic such as side-road entries, a spatial gating threshold $\epsilon_{lat}$ is applied; if the minimum displacement exceeds this limit, the match is rejected, and $\mathcal{T}_b$ is instantiated as a new global identity.
\end{itemize}

To handle stale tracklets, we incorporate a static Time-to-Live (TTL) mechanism. Entries in $\mathcal{B}$ are invalidated if they exceed a temporal threshold $\epsilon_{time}$, preventing stale associations.

\begin{algorithm}[t]
\caption{Topology-Aware Spatiotemporal Handover}
\label{alg:handover_logic}
\begin{algorithmic}[1]
\Require Streams $S_{1\dots N}$, Graph $\mathcal{G}$, Polygons $\mathcal{R}_{ovlp}$, Search Window $\Delta t$
\Ensure Unified Global Trajectories $\mathcal{T}_{global}$

\State $\mathcal{B} \gets \text{InitBuffers}(\mathcal{G})$ \Comment{Directional Metadata Buffers}
\State $gID\_counter \gets 0$

\While{System is Running}
    \For{$c \in \{1 \dots N\}$ \textbf{in parallel}} 
        \State $\mathcal{T}^L \gets \text{LocalPerception}(S_c)$ 
        
        \For{\textbf{each} track $\tau \in \mathcal{T}^L$}
            \State $\mathbf{v}, \theta \gets \text{EstKinematics}(\tau)$
            \State $Z \gets \text{GetZone}(\tau.pos, \theta)$ \Comment{Eq. 8}

            \If{$\tau.gID = \emptyset$} \Comment{\textbf{Entry Logic}}
                % Modified Line 10: Explicitly showing temporal search for gaps
                \State $\tau_{match}, dist \gets \text{QueryMatch}(\mathcal{B}_{up \to c}^{Z}, \tau.y_{rel}, \Delta t)$ 
                
                \If{$\tau_{match} \neq \text{None} \land dist < \epsilon_{lat}$}
                    \State $\tau.gID \gets \tau_{match}.gID$
                    \State $\text{Remove}(\mathcal{B}_{up \to c}^{Z}, \tau_{match})$
                \Else
                    \State $\tau.gID \gets ++gID\_counter$
                \EndIf
            \EndIf

            \If{$\tau \in \mathcal{R}_{ovlp}$} \Comment{\textbf{Exit Logic}}
                \If{$\tau.gID \notin \mathcal{B}_{c \to next}^{Z}$}
                    \State $\text{Push}(\mathcal{B}_{c \to next}^{Z}, \{\tau.gID, \tau.y_{rel}\})$
                \EndIf
            \EndIf
            
            \State $\text{UpdateGlobalState}(\mathcal{T}_{global}, \tau)$
        \EndFor
    \EndFor
    \State $\text{CheckTimeouts}(\mathcal{B}, \epsilon_{time})$
\EndWhile
\end{algorithmic}
\end{algorithm}

\subsection{Kinematic State Estimation}
\label{subsec:Kinematics}
To support the directional filtering, raw pixel displacements are converted to metric velocities. We employ a GSD-based linear calibration coefficient $\lambda$ (pixel/m). To mitigate quantization noise, the instantaneous velocity $v_t$ is smoothed over a sliding window $k$:
\begin{equation}
    v_t = \frac{|| \mathbf{p}_t - \mathbf{p}_{t-k} ||_2 \cdot \lambda}{k \cdot \Delta t} \times 3.6 \quad [km/h]
\end{equation}
This metric velocity determines the vehicle's stopped or moving status, providing additional context for the global identity management.

\subsection{Temporal Synchronization and Scalable Architecture}
\label{subsec:System}
The validity of the topology-aware handover (Alg.~\ref{alg:handover_logic}) relies fundamentally on the assumption that the global state snapshot $\mathcal{S}_t$ represents a temporally aligned observation of the entire sensor network. In a distributed UAV constellation, asynchronous frame arrival times can induce temporal misalignments, potentially causing causality violations in the trajectory handover, such as a vehicle appearing in a downstream node before physically departing the upstream region.

To address this challenge, we propose a Synchronized Parallel Execution Model that logically decouples the system architecture into two distinct strata to ensure both scalability and consistency. The first stratum functions as an Asynchronous Perception Layer, where local perception modules $\Phi_{loc}$ operate independently on each video stream $S_c$. This parallelization strategy ensures that the computational complexity of object detection scales linearly $\mathcal{O}(N)$ rather than exponentially with the number of UAV nodes, thereby preventing processing bottlenecks as the swarm size increases.

Complementing this, the second stratum acts as a Synchronous Global Consumer. We introduce a Global Synchronization Barrier to enforce strict temporal consistency across the distributed inputs. This mechanism buffers local tracklets $\mathcal{T}^L$ from all incoming streams until a unified timestamp $t$ is reached. The topological handover logic is executed only when the network-wide state is fully synchronized. This hierarchical design guarantees that spatiotemporal queue operations, specifically the Push and Pop events, are performed on a theoretically valid snapshot of the traffic flow, effectively isolating the computational intensity of visual inference from the lightweight topological reasoning.

\begin{figure}[t]
\centering
\includegraphics[width=\columnwidth]{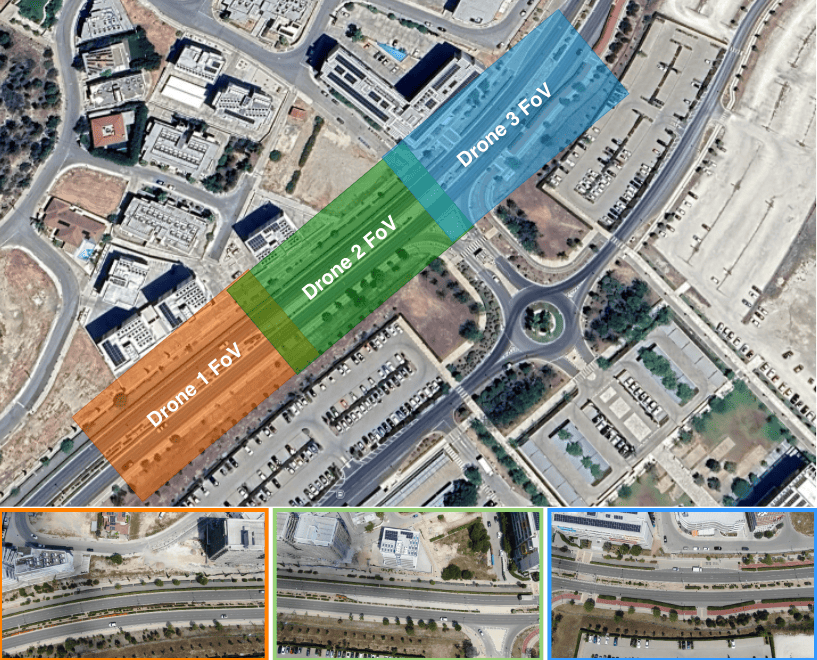}
\caption{Experimental data acquisition. \textbf{Top:} Map of the Aglantzia area showing the locations of the three monitored urban road segments and the corresponding monitoring footprints. \textbf{Bottom:} Representative 4K frames captured by a DJI Mavic 2 Enterprise drone over each monitored segment.}
\label{fig:data_acquisition}
\end{figure}

\begin{figure}[ht]
\centering
\includegraphics[width=\columnwidth]{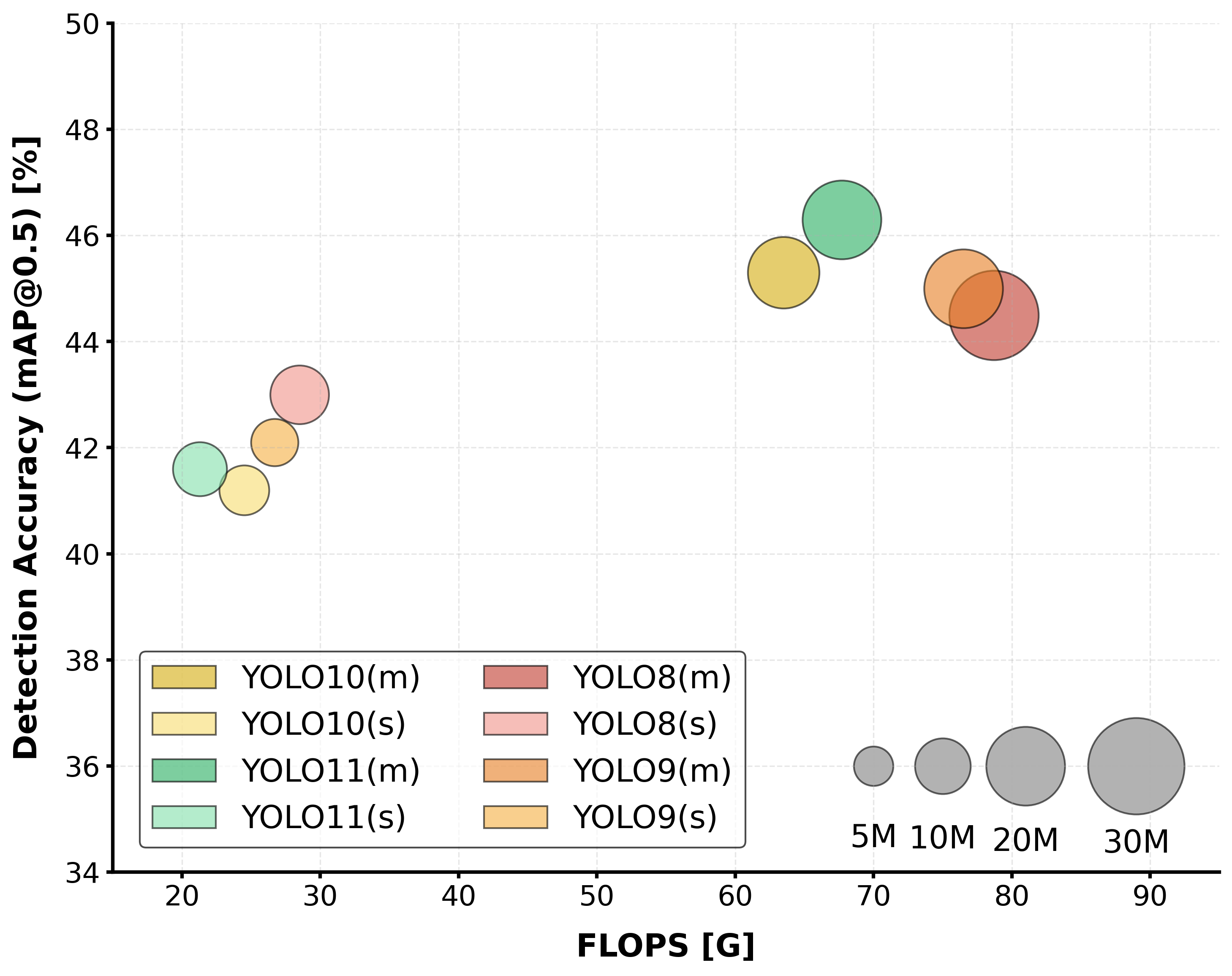}
\caption{Accuracy-efficiency trade-off analysis on the VisDrone2019-val dataset. The scatter plot illustrates the relationship between computational cost (GFLOPS) and detection accuracy (mAP@0.5) for various YOLO architectures. The area of each marker is proportional to the model's parameter count (model size), providing a comparison of model complexity against performance.}
\label{fig:visdrone_tradeoff}
\end{figure}
\section{Experimental Results and Analysis}
\label{sec:Evaluation}

We designed a comprehensive evaluation protocol to validate the two core claims of this work: (1) that topology-aware constraints outperform appearance-based Re-ID in nadir-view tracking, and (2) that the proposed pipeline is computationally efficient enough for edge deployment.

\begin{figure}[t]
\centering
\includegraphics[width=1.0\columnwidth]{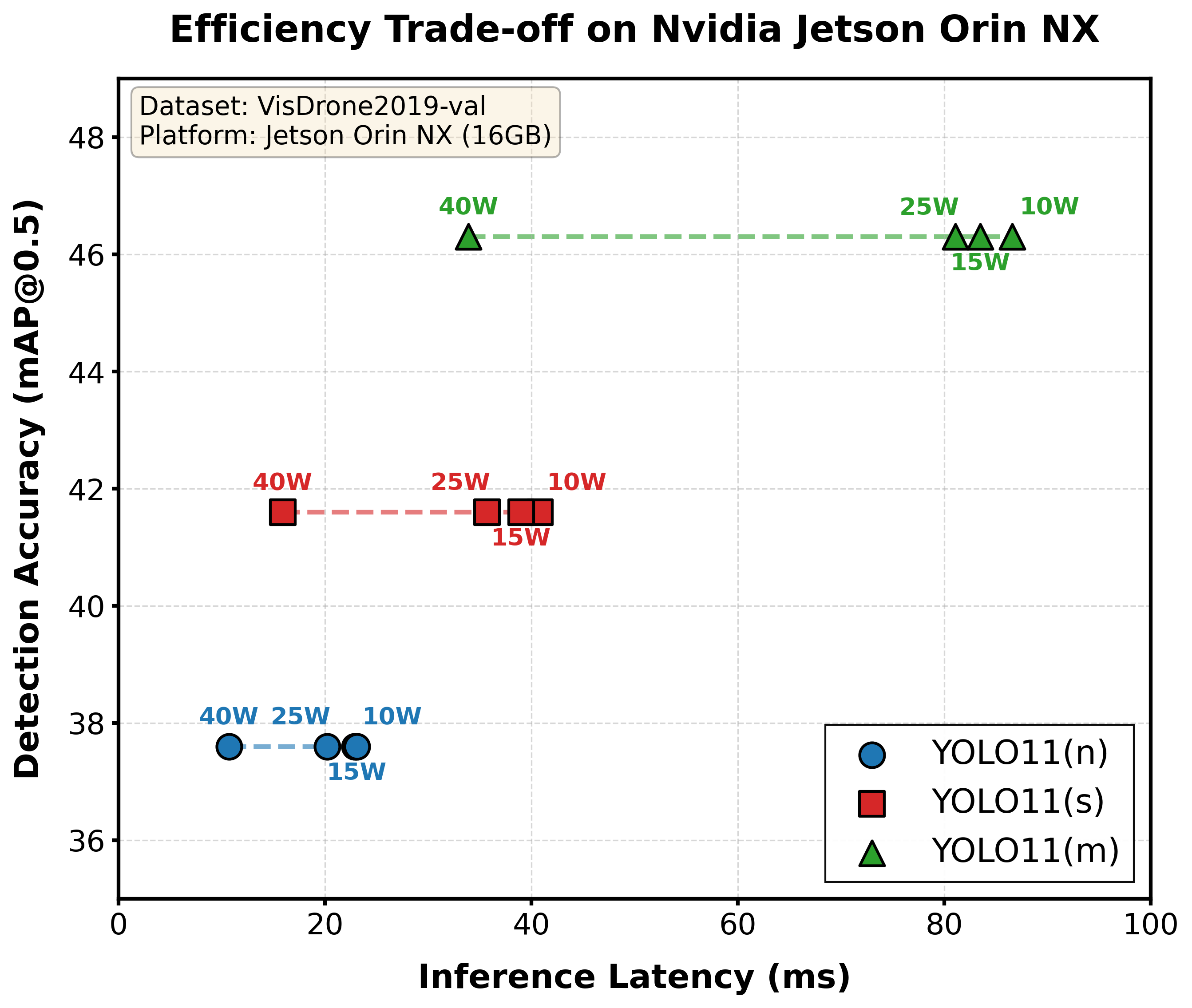} 
\caption{Inference Latency vs. Accuracy for YOLO11 variants on Jetson Orin NX (15W Mode). The YOLO11s model (Red Star) is selected as the optimal operating point.}
\label{fig:efficiency}
\end{figure}

\subsection{Experimental Setup}
\label{subsec:Setup}

\subsubsection{Dataset and Scenarios}
We curated a custom multi-UAV dataset, captured explicitly in a nadir-view (top-down) configuration. Existing benchmarks predominantly feature oblique, ground-level CCTV views, which lack the scale variation and specific "roof-only" visual ambiguity characteristic of aerial monitoring.

The dataset covers a 500m urban corridor comprising three synchronized 4K streams. Notably, the topology is not merely a closed highway; the segment between $UAV_2$ and $UAV_3$ includes a complex intersection feeding a university campus entrance. The monitoring footprints and representative 4K frames are presented in Fig.~\ref{fig:data_acquisition}. This introduces realistic "Merge and Diverge" behaviors, where vehicles enter the tracking stream from side roads or exit mid-chain.

To systematically evaluate robustness, we categorized the footage into three distinct traffic regimes: Set A (Free-Flow), characterized by low density ($<10$ veh/km) and high speeds ($>50$ km/h); Set B (Congestion), featuring high density ($>50$ veh/km) with stop-and-go waves; and Set C (Overtaking), involving complex interactions with frequent lane changes. Finally, to ensure rigorous validation, the ground truth trajectories and global identities were manually annotated using the CVAT tool at 10 FPS intervals.

\subsubsection{Evaluation Metrics}
To rigorously quantify global identity persistence, we adopt standard MOT metrics alongside specific indicators: HOSR (Handover Success Rate) measures the ratio of correct global associations in overlap zones; IDF1 quantifies global identity stability against ground truth; and FPS evaluates end-to-end throughput including detection and handover logic.

\subsubsection{Implementation Details}
Experiments were conducted on two platforms: a Server Node (NVIDIA V100 GPU) for baseline comparison and an Edge Node (NVIDIA Jetson Orin NX 16GB) for feasibility testing. The detector (YOLO11s) was trained on the VisDrone2019 dataset~\cite{du2019visdrone} for 300 epochs, achieving a baseline mAP@0.5 of 41.6\% (see Fig.~\ref{fig:visdrone_tradeoff}). 

\subsection{Comparative Analysis}
We benchmark our Topology-Aware approach against leading multi-camera tracking paradigms. To rigorously evaluate the algorithmic throughput limit independent of onboard hardware constraints, the comparative experiments in this section were conducted on a server node equipped with an NVIDIA Tesla V100 GPU. As shown in Table~\ref{tab:sota_comparison}, we compare against:

\subsubsection{DeepSORT (Re-ID)} Standard Kalman filter integration with appearance embedding matching for association~\cite{Wojke2018deep}.
\subsubsection{FastReID (Re-ID)} A heavy-weight Re-ID model utilizing a ResNet50 backbone for post-hoc trajectory association~\cite{he2023fastreid}.
\subsubsection{ByteTrack (Motion)} A pure motion-based tracker serving as a naive baseline without topological constraints~\cite{zhang2022bytetrack}.

\begin{figure*}[t]
\centering
\includegraphics[width=\textwidth]{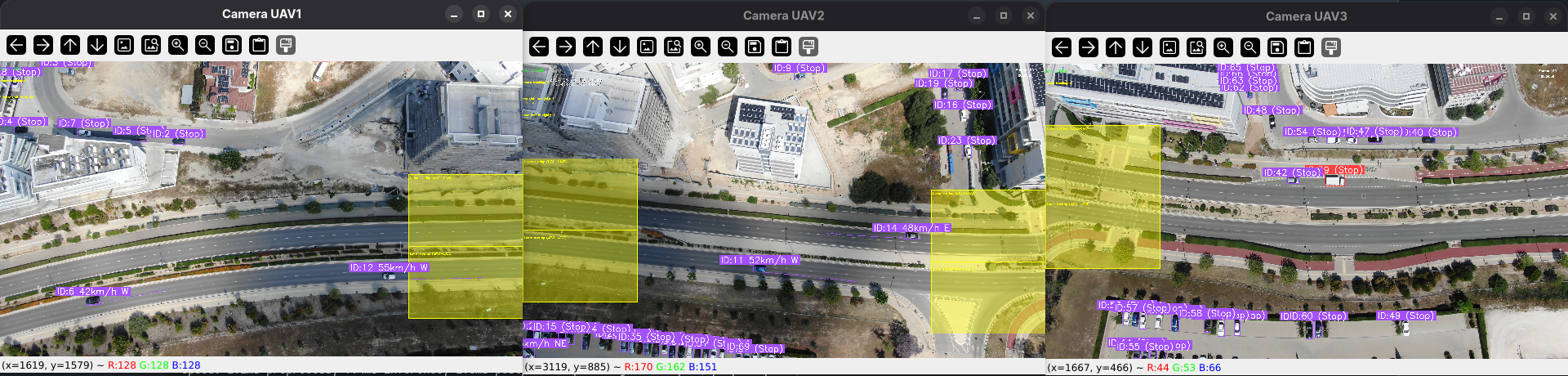}
\caption{Operational snapshot of the proposed Unified Aerial Surveillance system. Three synchronized UAV video streams $UAV_{1}$ to $UAV_{3}$ are annotated in real time with kinematic state estimates, including the persistent global identity $ID_G$, metric velocity, heading, and motion status. Yellow regions denote the calibrated overlap polygons $\mathcal{P}_{\text{overlap}}$ where the spatiotemporal queue performs identity handover between adjacent fields of view.}
\label{fig:operational_snapshot}
\end{figure*}

\begin{table}[ht]
\centering
\caption{Comparative Analysis on the UAV-Chain Dataset.}
\label{tab:sota_comparison}
\resizebox{\columnwidth}{!}{%
\begin{tabular}{@{}lcccc@{}}
\toprule
\textbf{Method} & \textbf{Mechanism} & \textbf{HOSR (\%)} & \textbf{IDF1 (\%)} & \textbf{FPS (Server)} \\ \midrule
ByteTrack (Naive) & Motion IoU & 12.4 & 34.2 & \textbf{65.2} \\
DeepSORT & Re-ID (CNN) & 68.3 & 55.1 & 22.8 \\
FastReID (ResNet) & Re-ID (Large) & 74.1 & 62.8 & 8.4 \\
\midrule
\textbf{Ours (Proposed)} & \textbf{Topology + Meta} & \textbf{99.8} & \textbf{96.5} & 62.1 \\ \bottomrule
\end{tabular}%
}
\end{table}

The appearance-based methods (DeepSORT, FastReID) suffer significantly from "nadir-view ambiguity." FastReID, despite its heavy computational cost (dropping server-side FPS to 8.4), only achieves 74.1\% HOSR because visually identical vehicles (e.g., white sedans) are indistinguishable in aerial views. In contrast, our approach leverages deterministic topological constraints, achieving near-perfect handover (99.8\%) while maintaining real-time performance (62.1 FPS) comparable to the lightweight ByteTrack baseline. This confirms that for structured aerial surveillance, a vehicle's spatiotemporal context often matters more than its visual appearance.

\begin{table}[ht]
\centering
\caption{System Robustness Analysis Across Traffic Regimes}
\label{tab:scenario_analysis}
\resizebox{\columnwidth}{!}{%
\begin{tabular}{@{}lccc@{}}
\toprule
\textbf{Regime} & \textbf{Primary Challenge} & \textbf{HOSR} & \textbf{Latency (ms)} \\ \midrule
Set A (Free-Flow) & High Velocity Blur & 99.8\% & 14.2 \\
Set B (Congestion) & Queue Timeout & 92.4\% & 18.5 \\
Set C (Overtaking) & Parallel Handovers & 98.6\% & 15.1 \\ \bottomrule
\end{tabular}%
}
\end{table}

\subsection{Robustness Analysis Across Traffic Regimes}
To validate the reliability of the queue-based logic, we analyzed performance across different traffic scenarios (Table~\ref{tab:scenario_analysis}). In Free-Flow (Set A), we performed a sensitivity analysis regarding topological noise. The system maintained an HOSR $>98\%$ even when momentary UAV drifts caused estimated overlap misalignments of up to 15--20 meters (approximately 30\% of the overlap length). The metadata matching window ($\epsilon_{lat}$) successfully compensated for these geometric errors, confirming that strict pixel-perfect calibration is not a prerequisite for successful handover.

The Lateral-Aware Spatial Matching logic proved critical in Overtaking (Set C) scenarios, successfully resolving 98.6\% of parallel handover events where simple FIFO logic would have failed. To quantify this, a baseline ablation was conducted by disabling the lateral-aware matching; the HOSR in Set C dropped to 74.5\%, confirming that spatiotemporal metadata is indispensable for resolving parallel maneuvers. A slight performance drop is observed in Congestion (Set B), where prolonged stops occasionally exceeded the static TTL threshold, causing identity resets. This suggests a future direction for adaptive TTL based on estimated traffic flow velocity.

\subsection{Edge Deployment Feasibility}
\label{subsec:Edge_Deployment}
While the server-side evaluation (Table~\ref{tab:sota_comparison}) validates the logical correctness and maximum throughput potential of the proposed pipeline, practical UAV operations necessitate onboard processing under strict power constraints. To assess the system's real-world viability, we transferred the implementation to an NVIDIA Jetson Orin NX (16GB) platform, a representative embedded platform for aerial robotics, to evaluate energy-efficiency trade-offs.

As illustrated in Fig.~\ref{fig:efficiency}, we analyzed the YOLO11 detector family across different power modes. The YOLO11s model emerged as the balanced candidate for this specific hardware, achieving 25.6 FPS in 15W Mode (approximately 0.58 J/frame). Although the nano variant (YOLO11n) offered higher throughput (43 FPS), its reduced recall for small objects (mAP 37.6 vs 41.6) was observed to negatively impact the tracklet continuity essential for the handover trigger. Conversely, YOLO11m provided marginal accuracy gains but introduced latency that exceeded our real-time budget.

Crucially, this evaluation validates a decentralized processing paradigm, where the reported 25.6 FPS represents the dedicated throughput for an individual video stream. This result confirms that within a fully autonomous swarm in which each UAV is equipped with a dedicated onboard compute unit, the system can independently sustain real-time perception. Such an architecture effectively eliminates the latency and bandwidth bottlenecks typically associated with transmitting multiple high-resolution feeds to a centralized edge server, thereby satisfying the operational requirements for low-latency distributed tracking. Furthermore, the hardware-in-the-loop profile on the NVIDIA Orin NX serving as the representative compute payload utilized in industrial UAV swarms, this confirms the system's readiness for on-board UAV integration.

\subsection{Qualitative Analysis}
Figure~\ref{fig:operational_snapshot} visualizes the operational dashboard of the Unified Aerial Surveillance system, verifying its capability to process synchronized high-resolution streams from a distributed UAV constellation. This visualization provides qualitative validation on four key dimensions: 1) Concurrent Multi-Stream Processing: The interface demonstrates throughput capacity by rendering three simultaneous 4K feeds ($UAV_{1}\rightarrow UAV_{3}$), validating the efficiency of the parallelized architecture. 2) Visualization of Topological Constraints: Geometric Overlap Polygons (yellow regions) are highlighted not merely as markers but as active spatial triggers. Their alignment confirms calibration effectiveness. 3) Granular Kinematic Estimation: The system maintains a unified identity namespace; vehicles in downstream nodes carry global IDs propagated from upstream, transforming isolated feeds into a cohesive sensor network. 4) Robustness to Topological Noise: Crucially, the system exhibits resilience to alignment errors. Unlike pixel-level stitching, our metadata-driven logic ($\mathbf{M}=\{v,\theta,y_{rel}\}$) allows for "Soft Handovers," re-associating identities even across unmonitored gaps caused by wind-induced UAV drift.
\section{Conclusion and Future Work}
\label{sec:Conclusion}

This paper presents a unified MCMT framework designed to bridge the gap between isolated aerial observations and network-level traffic intelligence. By shifting the paradigm from fragile appearance-based Re-ID to deterministic topological reasoning, we successfully address the "nadir-view ambiguity" that fundamentally limits traditional tracking pipelines in aerial highway monitoring. Our experimental results demonstrate that exploiting the physical constraints of the roadway, particularly directionality and sequential flow, enables high-precision identity persistence (99.8\% HOSR) even in the presence of identical vehicle platoons. Furthermore, the successful integration of this logic into a high-throughput parallel pipeline verifies that sophisticated global association is achievable in real-time on resource-constrained edge hardware.

While our current implementation operates in the pixel domain to maximize speed, the ultimate frontier for ITS lies in the convergence of topological connectivity and geodetic precision. We acknowledge that our linear GSD factor assumes a strictly nadir view, necessitating homography-based calibration in future iterations to compensate for camera pitch. Furthermore, while resolving visual ambiguity, our deterministic logic remains sensitive to measurement noise; we aim to address cases exceeding the soft handover window’s capacity through probabilistic multi-modal association. Such a hybrid framework, coupled with georeferencing and stabilization, would provide a comprehensive solution for city-scale analytics, neutralizing environmental perturbations and trajectory fragmentation.

Future research will focus on three trajectories to enhance system autonomy. First, to address the limitations observed in congested scenarios, we aim to implement adaptive temporal gating, where queue timeout thresholds dynamically adjust based on real-time optical flow velocity estimates. Second, we plan to automate the calibration of handover zones. Currently, these regions require manual annotation; we intend to leverage image stitching algorithms to automatically detect and spatially align the overlapping fields of view, enabling rapid plug-and-play deployment. Finally, building on our edge deployment validation, we intend to investigate decentralized UAV-to-UAV communication. This would enable a fully autonomous swarm where identity handovers occur via direct peer-to-peer data links in the sky, removing the dependency on high-bandwidth transmission to a central ground station.
\section*{Acknowledgments}
This work was supported by the European UCPM program under grant agreement No 101193719 (COLLARIS2). It was also partially supported by the European Union's Horizon Europe research and innovation program under grant agreement No 101187121 (EUSOME) and from the Republic of Cyprus through the Deputy Ministry of Research, Innovation and Digital Policy.

\flushbottom
\balance

\bibliographystyle{IEEEtran}
\bibliography{references} 

\end{document}